\documentclass[letterpaper, 10 pt, conference]{ieeeconf}  

\IEEEoverridecommandlockouts

\usepackage{pstricks}
\usepackage{amsmath,amsfonts}
\usepackage{algorithmic}
\usepackage{algorithm}
\usepackage{array}
\usepackage[caption=false,font=normalsize,labelfont=sf,textfont=sf]{subfig}
\usepackage{textcomp}
\usepackage{stfloats}
\usepackage{url}
\usepackage{verbatim}
\usepackage{graphicx}
\usepackage{cite}
\usepackage[switch]{lineno}

\usepackage{orcidlink}
\usepackage{dsfont}
\usepackage{tabularx}
\usepackage{comment}

\usepackage{pstool}
\usepackage{mathtools}
\usepackage{color}
\usepackage{psfrag}
\usepackage{caption}

\usepackage{pdfpages}
\usepackage{soul}

\usepackage{pifont}
\usepackage{url}

\usepackage{graphicx}
\usepackage{pstricks}
\usepackage{verbatim}
\usepackage{psfrag}
\usepackage{pstricks}

\usepackage{graphicx}
\usepackage{longtable}
\usepackage{booktabs}
\usepackage{lscape}
\usepackage{array}
\usepackage{multirow}
\usepackage{bm}

\renewcommand{\keywords}[1]{\textbf{\textit{Index terms---}} #1}

 \newtheorem{Optimi}{\bf Optimization Problem}

\newcommand{\ds}{\displaystyle}

 \newcommand{\mat}[2]{\left[\begin{array}{#1} #2 \end{array}\right]}

   \def\p{{\bf p}}
\def\q{{\bf q}}

 \def\B{{\bf B}}  
  \def\G{{\bf G}} 
\def\I{{\bf I}} \def\J{{\bf J}}  
\def\M{{\bf M}}   
   
  \def\W{{\bf W}}

 \newtheorem{Prop}{\bf Property}

\newcommand\notsotiny{\@setfontsize\notsotiny\@vipt\@viipt}
 \makeatother

\title{\LARGE \bf
A Minimum-Energy Control Approach for Redundant Mobile Manipulators in Physical Human-Robot Interaction Applications 
}

\author{Davide Tebaldi$^{1}$\orcidlink{0000-0003-1432-0489}, Niccolò Paradisi$^{2}$\orcidlink{0009-0001-9903-8484}, Fabio Pini$^{1}$\orcidlink{0000-0001-9263-426X} and Luigi Biagiotti$^{1}$\orcidlink{0000-0002-2343-6929}
\thanks{$^{1}$The authors are with the ``Enzo Ferrari'' Department of Engineering, University of Modena and Reggio Emilia, via Pietro Vivarelli 10, 41125 Modena, Italy, e-mail: davide.tebaldi@unimore.it, fabio.pini@unimore.it, luigi.biagiotti@unimore.it. $^{2}$The author is with the Department of Engineering for Innovation Medicine, University of Verona, e-mail: \newline niccolo.paradisi@studenti.univr.it.}
\thanks{The work was funded under the National Recovery and Resilience Plan (NRRP), Mission 04 Component 2 Investment 1.5 – NextGenerationEU, Call for tender n. 3277 dated 30/12/2021 Award Number:  0001052 dated 23/06/2022, and was supported by the Italian National Recovery and Resilience Plan (PNRR), Mission 4 “Education and Research”, Component C2, Investment 1.1 “PRIN – Projects of Relevant National Interest”, Project I-SHARM: Intelligent SHared Autonomy for Robotic Manipulation Systems, Project ID 2022NTZRFM, CUP E53C24002600006.}
\thanks{© 20XX IEEE.  Personal use of this material is permitted.  Permission from IEEE must be obtained for all other uses, in any current or future media, including reprinting/republishing this material for advertising or promotional purposes, creating new collective works, for resale or redistribution to servers or lists, or reuse of any copyrighted component of this work in other works.}
}

\begin{document}

\maketitle
\thispagestyle{empty}
\pagestyle{empty}

\begin{abstract}
Research on mobile manipulation systems that physically interact with humans has expanded rapidly in recent years, opening  the way to tasks which could not be performed using fixed-base manipulators.
Within this context, developing suitable control methodologies is essential since mobile manipulators introduce additional degrees of freedom, making the design of control approaches more challenging and more prone to performance optimization.\\
This paper proposes a control approach
for a mobile manipulator, composed of a mobile base equipped
with a robotic arm mounted on the top, with the objective of
minimizing the overall kinetic energy stored in the whole-body mobile manipulator in physical human-robot interaction applications. The
approach is experimentally tested with reference to a peg-in-hole task, and the results demonstrate that the proposed
approach reduces the overall kinetic energy stored in the whole-body robotic system and improves the system performance compared with the benchmark method.
\end{abstract}



\section{INTRODUCTION}
\begin{figure}[t]
    \centering
    \includegraphics[clip,width=0.75\columnwidth]{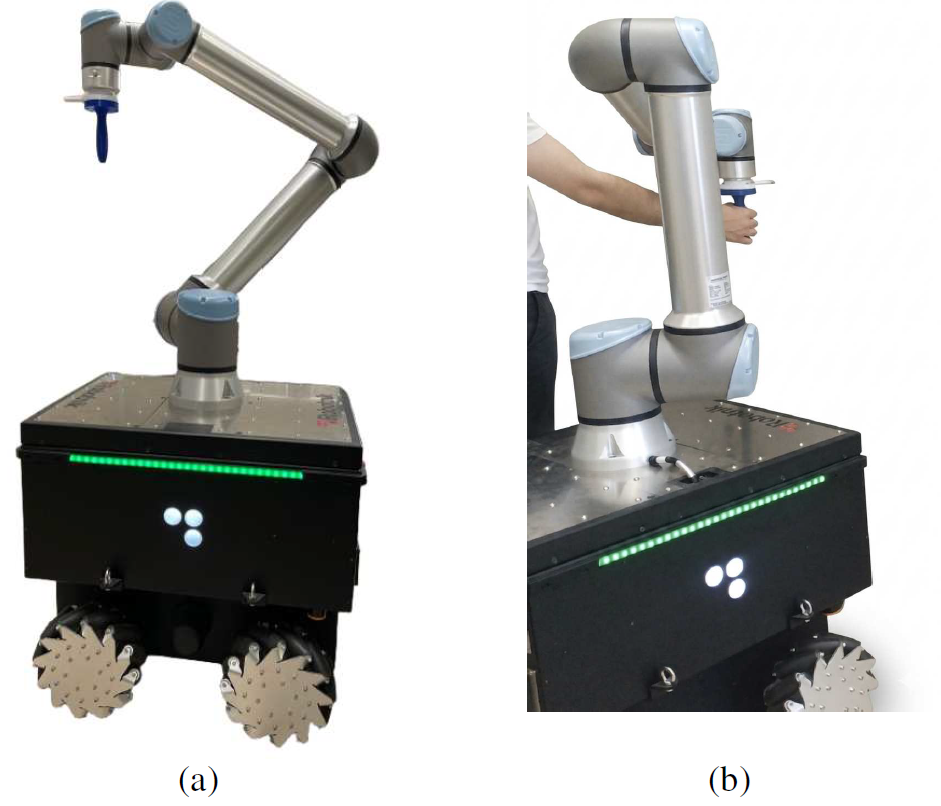}
\caption{(a) Considered mobile manipulator and (b) an example of Human/Mobile Manipulator cooperative task.}\label{fig:Mobile_System}
\vspace{-1.48mm}
 \end{figure}

\begin{figure}[t]
    \centering
    \vspace{2mm}
    \includegraphics[width=0.84\columnwidth]{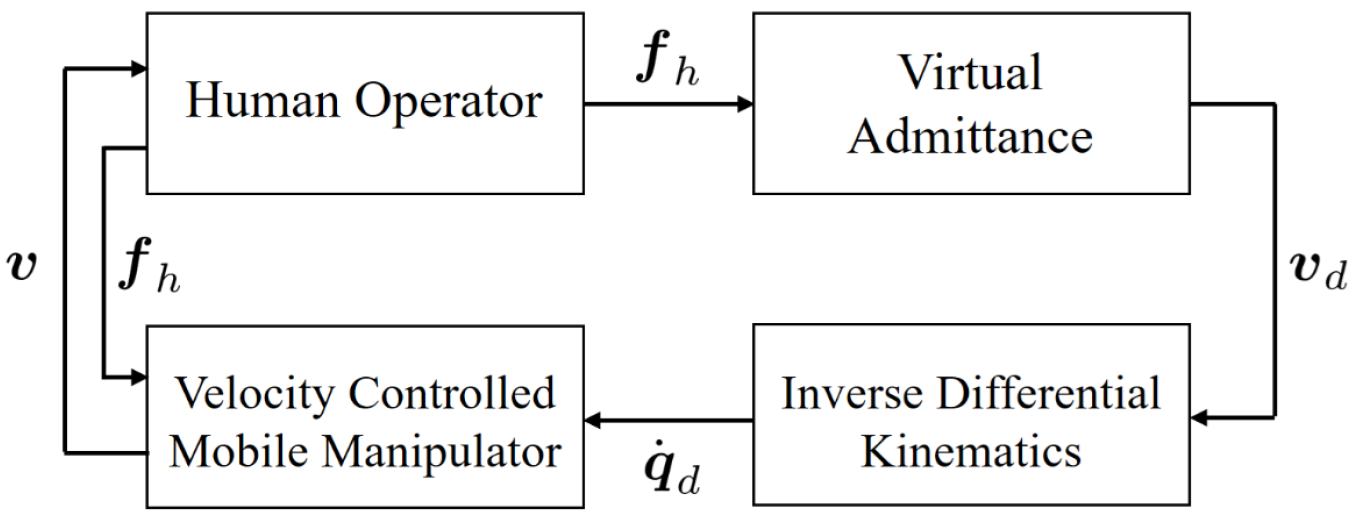}
    \vspace{2mm}
    \caption{Architecture for physical interaction between a human operator and a mobile manipulator through a virtual admittance model.}
\label{fig:archit_RL}
\end{figure}

\begin{figure}[t]
\centering
    \centering
   \vspace{-4.02mm}
    \includegraphics[width=0.68\columnwidth]{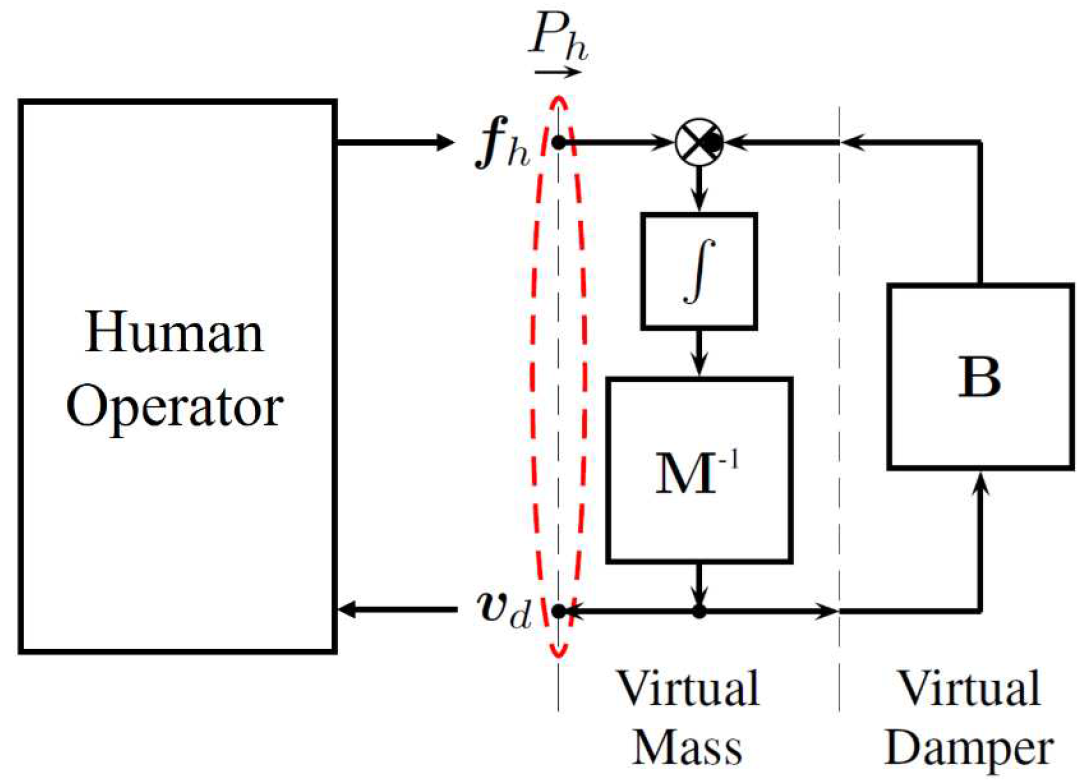}
 \caption{Power-Oriented Graph (POG) representation of the virtual admittance model, showing the energetic port of the interaction between the human operator and the admittance model to generate the desired end-effector twist $\boldsymbol{v}_d$.}\label{admittance_POG}
\end{figure}

The interest in collaborative robotics applications has significantly increased in the last two decades. This includes applications such as rehabilitation robotics \cite{DESHPANDE2022100371,7856889}, robot-assisted surgery \cite{8392699}, and industrial tasks \cite{ANGLERAUD2024102663}.\\
The aforementioned applications typically involve the use of fixed-base manipulators to accomplish these tasks. However, certain tasks such as co-transportation \cite{9548781} and others, which can be physically demanding and affect the operator's health, can gain significant benefit from the use of a mobile manipulator. This concept falls within the area of the so-called human augmentation field, which exploits the use of technologies in conjunction with human capabilities~\cite{Ganavi2023}. This makes it possible to overcome human body limitations and to enhance human abilities.\\
The considered mobile manipulator is shown in Fig.~\ref{fig:Mobile_System}(a), while the considered scenario is depicted in Fig.~\ref{fig:Mobile_System}(b), showing the mobile robot equipped with a handle. In this scenario, the human interacts with the end-effector to perform a peg-in-hole task. A classical architecture for human-robot interaction applications is shown in Fig.~\ref{fig:archit_RL}~\cite{9197115,9812225,9812501,GIAMMARINO2024103240,SIRINTUNA2024104725}. This architecture depicts a human operator interacting with the robot through the following virtual admittance dynamics
\begin{equation}\label{atmitt_model}
\M \dot{\boldsymbol{v}}_d + \B \boldsymbol{v}_d = \boldsymbol{f}_h,
\end{equation}
where $\M, \B \in \mathbb{R}^{(6 \times 6)}$ are the virtual mass and damper matrices, $\boldsymbol{f}_h \in \mathbb{R}^{(6 \times 1)}$ is the human wrench, and $\boldsymbol{v}_d \in \mathbb{R}^{(6 \times 1)}$ is the generated desired twist in the operational space of the end-effector. In Fig.~\ref{admittance_POG}, the admittance model~\eqref{atmitt_model} is represented using the Power-Oriented Graphs (POG) formalism \cite{Tebaldi2025}, which clearly highlights the energetic port through which the human operator interacts with the virtual admittance model to generate the desired end-effector twist~$\boldsymbol{v}_d$.

By solving the Inverse Differential Kinematics (IDK) in Fig.~\ref{fig:archit_RL}, the desired velocities $\dot{\q}_d$ in the joint space for the mobile base and for the manipulator arm are determined.

In this paper, we propose an approach  to solve the IDK problem which enables to minimize the kinetic energy stored in the whole-body robotic system.  

The advantages of the proposed control, emerging from the experimental validation, include a smoother user experience compared to the considered benchmark, as well as the expected reduction of the overall kinetic energy stored in the whole-body robotic system. Minimizing the kinetic energy stored in the system also translates into minimizing the movements of the heaviest system component, that is the mobile base weighing 115 kg. In turn, this also allows to enhance the safety of the resulting physical interaction between the human and the mobile manipulator.

The remainder of this paper is organized as follows. The related works are discussed in Sec.~\ref{Related_Works_sect}. The proposed control approach is presented in Sec.~\ref{prop_appr}, while the experimental setup is described in Sec.~\ref{experimental_results}. The obtained results are presented and discussed in Sec.~\ref{res_pres_and_disc}, while the conclusions of this work are reported in Sec.~\ref{concl_sect}.

\section{Related Works}\label{Related_Works_sect}
Most of the related works in the literature employ standard inverse differential kinematics or inverse kinematics approaches for redundant robots to transform velocities from the operational space to the joint space.
For instance, in \cite{9261373} they propose a control approach based on inverse kinematics. Specifically, they apply the FABRIK algorithm \cite{aristidou2009inverse}, originally meant for fixed-base manipulators, to mobile manipulators, with the objective of performing path tracking tasks. This approach allows to drive mobile base and manipulator arm  while satisfying additional requirements, such as obstacle avoidance and joint limits avoidance.
In \cite{ZHANG2023105273}, the inverse kinematics problem is solved as an optimization problem, to avoid joint limits and perform collision avoidance.

In the control framework illustrated in Fig.~\ref{fig:archit_RL}, the inverse kinematics of the mobile manipulator is not needed. However, as previously mentioned, it is necessary to solve the inverse differential kinematics problem, which consists in solving the following relation with respect to $\dot{\q}_d$:\\[-6mm]

\begin{equation}\label{diff_kin_inv_first}
\boldsymbol{v}_{d} = \J(\q) \dot{\q}_{d}, \hspace{4mm} \mbox{where} \hspace{4mm} \J(\q)=\begin{bmatrix}
\J_a(\q_{a}) & \J_b
\end{bmatrix}
\end{equation}
is the Jacobian of the whole system, whereas
$\J_a \!\in \!\mathbb{R}^{6\times 6}$ and $\J_b \!\in \!\mathbb{R}^{6\times 3}$ are the Jacobians of the manipulator arm and of the mobile base, respectively.

The desired end-effector twist in the operational space and the desired joint velocity vector are given by, respectively:
\begin{equation}\label{velocity_vectors}
\boldsymbol{v}_{d}\!=\!\mat{@{}c@{}}{v_{x_{d}} \\ v_{y_{d}} \\ v_{z_{d}} \\ \omega_{x_{d}} \\ \omega_{y_{d}} \\ \omega_{z_{d}}}\!\! \hspace{5.2mm} \mbox{and} \hspace{5.2mm}
\dot{\q}_{d}\!=\!\mat{@{}c@{}}{\dot{\q}_{a_d} \\ \dot{\q}_{b_d}
}\!=\!
\mat{@{}c@{}}{\dot{q}_{1_{d}} \\ \vdots \\
\dot{q}_{6_{d}} \\
v_{bx_{d}} \\
v_{by_{d}} \\
\omega_{bz_{d}}
}.
\end{equation}
Note that vector $\dot{\q}_d$ is composed of both the desired arm joint velocities $\dot{\q}_{a_d}$ and the desired virtual joint velocities $\dot{\q}_{b_d}$ of the mobile base, which must be transformed and applied to the base’s wheels. Since an omnidirectional base is considered, these virtual velocities correspond to the linear velocities $v_{bx_{d}}$, $v_{by_{d}}$ along the $x$, $y$ directions and to the angular velocity $\omega_{bz_{d}}$ about the $z$ axis.

In \cite{9695298}, the Inverse Differential Kinematics (IDK) is solved by formulating a Quadratic Programming (QP) problem, minimizing a joint velocities quadratic cost that includes a manipulability index of the robot arm. The optimization is performed by including the constraint indicated in Eq.~(\ref{diff_kin_inv_first}), so as to guarantee an admissible solution of the IDK and further constraints in terms of joint velocity and position limits. It is important to highlight that the method they propose is intended for general-purpose applications; however, due to the computational time required for solving the QP problem (which, as reported in the paper, takes at least 4.8 seconds) it is not suitable for real-time scenarios such as the one described in the architecture of Fig.~\ref{fig:archit_RL}, where the reference twist is generated through human–robot interaction and must be translated in real time into corresponding joint velocities. For this reason, in all works involving physical interaction between a mobile manipulator and a human operator, the inverse differential kinematics problem is typically addressed using a closed-form solution, generally based on the pseudoinverse of the Jacobian matrix $\J(\q)$ \cite{9197115,9812225,9812501}. For instance, this approach is adopted in \cite{biomimetics9020104}, where the authors consider a scenario in which a mobile manipulator assists individuals with mobility impairments in walking. In their method, virtual damping is adaptively tuned using fuzzy logic based on the human's intent, and a null-space controller is employed to avoid singular configurations.

The approach proposed in \cite{GIAMMARINO2024103240,SIRINTUNA2024104725}, which serves as a benchmark in the experimental evaluation of our method, addresses the IDK problem using a prioritized control technique.
 The resulting solution consists of two contributions, namely:
\begin{equation}\label{cons_solut}
\dot{\q}_{d}=\dot{\q}_{d,p}+\mathcal{N(\J(\q))}\dot{\q}_{d,s},
\end{equation}
where $\mathcal{N}(\J(\q)) = \I - \J^{\dagger}(\q)\J(\q)$ is a projector onto the null space of the Jacobian. 
The term $\dot{\q}_{d,p}$ accomplishes a primary task consisting in solving the damped least-squares problem minimizing the following objective function:
\begin{equation}\label{first_task}
\mathcal{L}_p(\dot \q_d) = \frac{1}{2}\|\boldsymbol{v}_{d} - \J(\q) \dot{\q}_{d} \|^2_{\W_\alpha} + \|\dot{\q}_{d} \|^2_{\W_\beta},
\end{equation}
with the two terms $\W_\alpha$ and $\W_\beta$ denoting two
weight matrices. The term $\dot{\q}_{d,s}$ accomplishes a secondary task consisting, in the considered case, in
minimizing the distance from a preferred configuration $\q_{des}$ \cite{GIAMMARINO2024103240}:
\begin{equation}\label{second_task}
\mathcal{L}_s(\q) = \frac{1}{2}\|\q_{des} - \q  \|^2_{\G},
\end{equation}
with the term $\G$ being a 
gain matrix.
The closed-form solution of \eqref{cons_solut} with the objective functions \eqref{first_task} and \eqref{second_task} is:
\begin{equation}\label{total_task}
\small
\begin{array}{r@{}c@{}l}
\dot{\q}_{d}&=&{(\J(\q)^{\!\top} \!\W_{\!\alpha} \J(\q)\!+\!\! \W_{\!\beta})}^{\!-\!1}\! \J(\q)^{\!\top} \! \W_{\!\alpha} \boldsymbol{v}_{d} \!+\!  \mathcal{N\!(\J(\q)\!)} \G (\q_{des}\!-\!\q).
\end{array}
\vspace{2mm}
\end{equation}

\section{Proposed Approach}\label{prop_appr}
Reference is made to the control architecture depicted in Fig.~\ref{fig:archit_RL}. The objective of the proposed approach is to solve the IDK problem so as to 
minimize the kinetic energy stored in the whole-body redundant mobile manipulator, which is composed of the mobile base and the robotic arm, during the execution of the task. In doing so, the movements of the heaviest system component, that is the mobile base, will be minimized, thereby enhancing safety in physical human-robot interaction. The resulting formulation corresponds to a weighted pseudoinverse using the inertia matrices $\M_a(\q_a)$ and $\M_b$ of the arm manipulator and of the mobile base, respectively.

\subsection{Optimization Problem Formulation}\label{prop_appr_optimi}
Reference is made to the redundant mobile manipulator consisting of a mobile base and a robotic arm shown in Fig.~\ref{fig:Mobile_System}. The IDK problem is solved through the following optimization problem.

\begin{Optimi}\label{opt:optim_pstar}
Find the optimal desired arm joint velocities $\dot{\q}^*_{a_d}$ and the optimal desired virtual joint velocities $\dot{\q}^*_{b_d}$ such that:
\begin{equation}
\min_{\dot{\q}_{a_d},\,\dot{\q}_{b_d}}E_K({\q}_{a},\dot{\q}_{a_d},\dot{\q}_{b_d})
    \label{eq:opt}
\end{equation}
where the objective function
\begin{equation}
    \label{tot_kin_en}
E_K({\q}_{a},\dot{\q}_{a_d},\dot{\q}_{b_d})=
    \frac{1}{2}\dot{\q}_{a_d}^\top \M_a(\q_{a})\,\dot{\q}_{a_d}
     + \frac{1}{2}\dot{\q}_{b_d}^\top \M_{b}\,\dot{\q}_{b_d}
\end{equation}
represents the total kinetic energy of the whole-body robotic system, with $\M_a(\q_a)$ and $\M_b$ being the inertia matrices of the arm manipulator and of the mobile base, respectively, subject to the constraint
\begin{equation}
    \J_a(\q_{a})\,\dot{\q}_{a_d} + \J_b\,\dot{\q}_{b_d}  = \boldsymbol{v}_d.
    \label{eq:task}
\end{equation}
\end{Optimi}
\vspace{1mm}
\noindent Note that the constraint \eqref{eq:task} derives from \eqref{diff_kin_inv_first} and \eqref{velocity_vectors}.

\begin{Prop}\label{opt:optim_cons}
The optimal solution $\dot{\q}_d^* = \begin{bmatrix} \dot{\q}_{a_d}^* & \dot{\q}_{b_d}^* \end{bmatrix}^\top$ solving the optimization problem \eqref{eq:opt} is given by
\begin{equation}\label{opti_solut}
    \dot{\q}_d^* = \J_{\M}^\dagger(\q)\,\boldsymbol{v}_d,
\end{equation}
where $\boldsymbol{v}_d$ is the desired operational-space twist defined in \eqref{velocity_vectors}, and $\q = \begin{bmatrix} \q_a & \q_b \end{bmatrix}^\top$ is the vector of generalized coordinates.
The matrix
\begin{equation}
 \J_{\M}^\dagger(\q) = \M(\q_a)^{-1}\J(\q)^\top
\big(\J(\q)\,\M(\q_a)^{-1}\J(\q)^\top\big)^{-1}
\end{equation}
is the dynamically consistent Jacobian pseudoinverse \cite{khatib2003unified}, with
the whole-body inertia matrix $\M(\q_a)$ defined as
\begin{equation}\label{matrix_M_q_a}    
    \M(\q_a) =
    \begin{bmatrix}
        \M_a(\q_a) & 0 \\
        0 & \M_b
    \end{bmatrix}.
\end{equation}
\vspace{0.2mm}
\end{Prop}
The off-diagonal inertial coupling terms between the arm and the mobile base are neglected in \eqref{matrix_M_q_a}. This approximation is motivated by the operating conditions of the considered HRI scenario, where the mobile base moves at deliberately low velocities and accelerations for safety reasons. Under these conditions, arm–base dynamic coupling effects are expected to be limited, especially compared with the dominant inertia of the mobile platform. Therefore, the block-diagonal approximation of the whole-body inertia matrix provides a sufficiently accurate model for computing the dynamically consistent pseudoinverse.\vspace{2mm}

{\it Proof.} The optimization problem~\eqref{eq:opt} can be equivalently rewritten in the following matrix form: 
\begin{equation}
    \min_{\dot{\q}_d}E_K(\q_{a},\,\dot{\q}_d),
    \quad \text{subject to} \quad
    \J(\q)\,\dot{\q}_d = \boldsymbol{v}_d,
        \label{eq:compact}
\end{equation}
where the total kinetic energy $E_K(\q_a,\,\dot{\q}_d)$ is expressed by the following matrix quadratic function:
\[E_K(\q_a,\,\dot{\q}_d)=    
    \frac{1}{2}\dot{\q}_d^\top \M(\q_a)\,\dot{\q}_d,
\]
with the inertia matrix $\M(\q_a)$ being defined in \eqref{matrix_M_q_a}.
The Lagrangian of~\eqref{eq:compact} is the following:
\[
\mathcal{L} = \frac{1}{2}\dot{\q}_d^\top \M(\q_a)\,\dot{\q}_d
+ \bm{\lambda}^\top ( \boldsymbol{v}_d - \J(\q)\dot{\q}_d ),
\]
which yields the following first-order optimality (KKT) conditions~\cite{goldstein1950classical}:  
\begin{align}
    \M(\q_a)\,\dot{\q}_d + \J(\q)^\top \bm{\lambda} &= 0,\label{eq:kkt1} \\
    \J(\q)\,\dot{\q}_d &= \boldsymbol{v}_d. \label{eq:kkt2}
\end{align}
Solving \eqref{eq:kkt1} and \eqref{eq:kkt2} for $\dot{\q}_d$ yields the optimal solution $\dot{\q}^*_d$ in \eqref{opti_solut}. $\hfill \Box$

From $\dot{\q}^*_d$, the optimal arm joint and base velocities $\dot{\q}^*_{a_d}$ and $\dot{\q}^*_{b_d}$ can be extracted recalling the definition of vector $\dot{\q}_d$ in \eqref{velocity_vectors}. 

The desired velocities $\dot{\q}_d$ in the joint space used in the control architecture of Fig.~\ref{fig:archit_RL} are then given by: 
\begin{equation}\label{cons_solut_1}
\dot{\q}_{d}=\dot{\q}^*_d+\mathcal{N(\J(\q))}\dot{\q}_{d,s},
\end{equation} 
where $\mathcal{N}(\J(\q)) = \I - \J_\M^{\dagger}(\q)\J(\q)$ is a projector onto the null space of the Jacobian and where 
\begin{equation}\label{cons_solut_2}
\dot{\q}_{d,s}=\G (\q_{des}-\q).
\end{equation} 
The purpose of the term \eqref{cons_solut_2} is to ensure the end-effector does not deviate excessively  from the preferred configuration, which is designed to ensure an ergonomic user configuration.

\section{Experimental Results}\label{experimental_results}

\begin{figure}[t]
    \centering
    \includegraphics[clip,width=0.59\columnwidth]{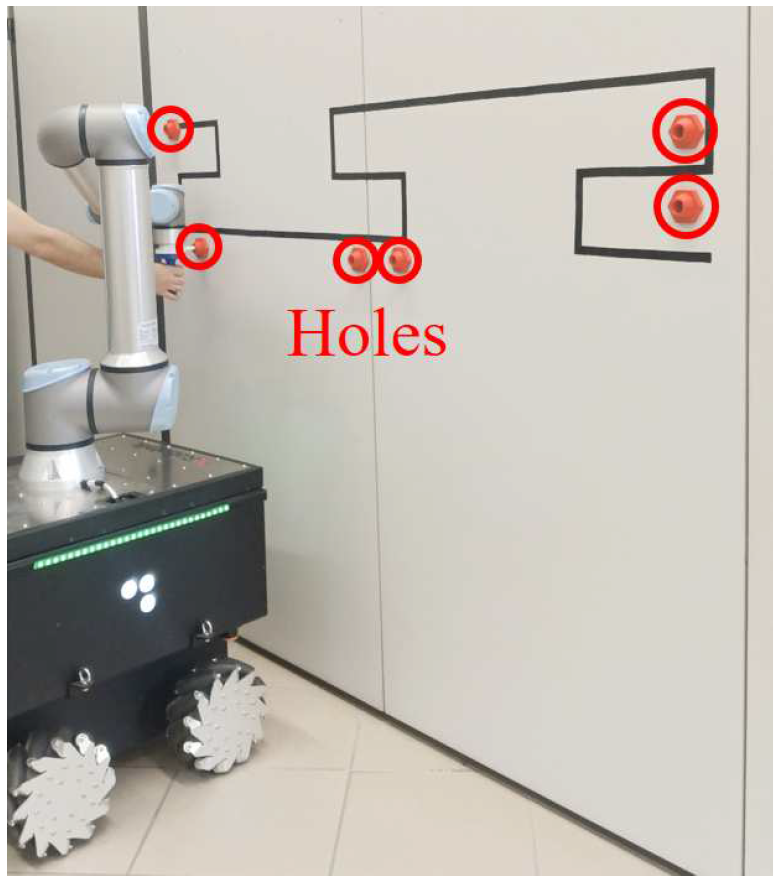}
\vspace{-1.68mm}
\caption{Experimental test scenario, consisting in performing a peg-in-hole task with respect to the holes encircled in red.}\label{fig:tasks}
\vspace{-1.48mm}
 \end{figure}

The employed mobile manipulator is shown in Fig.~\ref{fig:Mobile_System}, which is composed of the UR10e cobot and of the Robotnik RB-KAIROS+ omnidirectional mobile base. The force sensor used is an FTE-AXIA80. The end-effector of the arm consists in the peg-and-handle system shown in Fig.~\ref{fig:Mobile_System}. 

Twenty-seven participants took part in the study. The users were asked to perform the task depicted in Fig.~\ref{fig:tasks}, consisting in inserting the peg into a set of holes to perform a peg-in-hole task. The task was executed using the following three different control approaches: (a) the locomotion mode in \cite{GIAMMARINO2024103240,SIRINTUNA2024104725};
    (b) the switch mode in \cite{GIAMMARINO2024103240,SIRINTUNA2024104725}, where in the latter the user is allowed to switch between manipulation and locomotion operating modes whenever desired;
    (c) 
    the proposed minimum-energy approach, i.e. the control architecture of Fig.~\ref{fig:archit_RL} solving the IDK problem through
    \textbf{Optimization Problem 1} in Sec.~\ref{prop_appr_optimi}, where the inverse of matrix $\M(\q_a)$ in \eqref{matrix_M_q_a} has been computed using the \verb|inv| function of the submodule \verb|linalg| of the Python library \verb|NumPy|.
\begin{figure}[t]
 \centering
 \includegraphics[clip,width=0.959\linewidth]{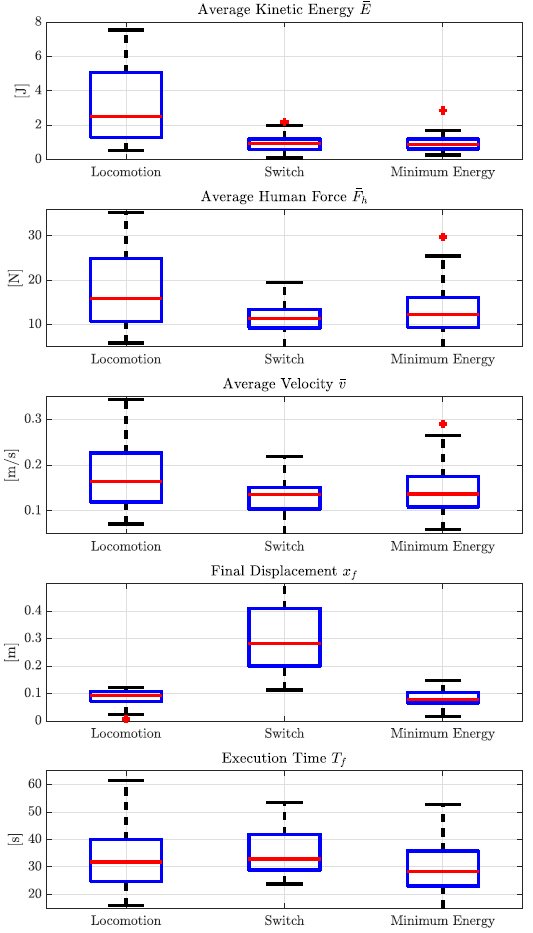}
    \vspace{-0.85mm}
    \caption{Average energy $\bar{E}$, average human force $\bar{F}_h$, average execution velocity $\bar{v}$, final displacement $x_f$ and execution time $T_f$ for the executions of the 27 users. 
    }
\label{fig:boxplots}
\vspace{-3.2mm}
 \end{figure}
The virtual inertia matrix $\M$ and the virtual friction matrix $\B$ in \eqref{atmitt_model} were set to $\M=\mathrm{diag}(4,\!\ldots,\!4)$ and $\B=\mathrm{diag}(75,\!\ldots,\!75)$, while the gain matrix $\G$, in \eqref{total_task} and \eqref{cons_solut_2} for the locomotion and for the minimum-energy approaches, respectively, was set to $\G=\mathrm{diag}(1,\!1,\!1,\!1,\!1,\!1,\!0,\!0,\!0)$. The acquired quantities include the norm of the velocity vector $\left\|\boldsymbol{v}(kT)\right\|$, the norm of the human force vector $\left\|\boldsymbol{f}_h(kT)\right\|$ (with the rotational components held constant due to the nature of the task, which requires the peg to remain orthogonal to the wall), and the total stored kinetic energy $E_K(kT)$, as defined in \eqref{tot_kin_en}, evaluated at discrete time instants $t = kT$. From these, the average energy $\bar{E}$, the average human force $\bar{F}_h$ and the average execution velocity $\bar{v}$ over the execution time $T_f$ have been computed as follows: 
\begin{equation}\label{altre_metrics}
\begin{array}{c}
\ds
\bar{E} = \frac{1}{N} \sum_{k=1}^{N} E_K(kT), \hspace{5.9mm}
\bar{F}_h = \frac{1}{N} \sum_{k=1}^{N} \left\|\boldsymbol{f}_h(kT)\right\|, \\[3.32mm]
\ds \bar{v} = \frac{1}{N} \sum_{k=1}^{N} \left\|\boldsymbol{v}(kT)\right\|, \hspace{5.9mm} \mbox{where} \;\; N = T_f/T.
\end{array}
\end{equation}
Let $\p_{des}=\mbox{FK}(\q_{des})$ be the desired position in the operational space, where ``FK'' is the forward kinematics operator and $\q_{des}$ is the reference position in the joint space introduced in \eqref{total_task} and \eqref{cons_solut_2}, and let $\p=\mbox{FK}(\q)$.
The final displacement $x_f$ is given by:
\begin{equation}\label{f_displ}  
x_f=\left\|\p(T_f)-\p_{des}
\right\|.
\end{equation}

The average energy $\bar{E}$, the average human force $\bar{F}_h$, the average execution velocity $\bar{v}$, the final displacement $x_f$, and the execution time $T_f$ have been acquired for each execution of the 27 different users for each of the three aforementioned control approaches: locomotion (a), switch (b) and the proposed minimum-energy approach (c).

The experimental results are shown in Fig.~\ref{fig:boxplots}, Fig.~\ref{figure_speed}, Fig.~\ref{figure_energy} as well as in Table~\ref{loco_appr}, Table~\ref{switch_appr} and Table~\ref{min_e_appr}, and are presented and discussed in the following section.

\subsection{Results Presentation and Discussion}\label{res_pres_and_disc}

The boxplots of Fig.~\ref{fig:boxplots} show the execution results of the 27 different users using the three considered control approaches, while the median values are reported in Table~\ref{loco_appr}, Table~\ref{switch_appr} and Table~\ref{min_e_appr}.
A Friedman test was conducted and, subsequently, a Wilcoxon signed-rank test followed by Holm-Bonferroni correction, to assess the statistical significance of the results of the different metrics between each pair of 
the three control approaches (a), (b) and (c) in Sec.~\ref{experimental_results}. The results are shown in Table~\ref{tab:posthoc_results}.

\begin{table}[t!]
    \centering
    \begin{tabular}{@{}c@{\;\;\;\;\;\;\;\;\;\;\;\;\;\;}c@{\;\;\;\;\;\;\;\;\;\;\;\;\;\;}c@{\;\;\;\;\;\;\;\;\;\;\;\;\;\;}c@{\;\;\;\;\;\;\;\;\;\;\;\;\;\;}c@{}}
\toprule
        $\bar{E}\!$ [J] & $\bar{F}_h\!$ [N] & $\bar{v}\!$ $\left[\mbox{m}/\mbox{s}\right]$ & $x_f$ $\!$ [m] &  $T_f$ $\!$ [s] \\
\midrule
$2.52$ & $15.84$ & $0.16$ & $0.09$ & $31.77$ \\
\bottomrule
    \end{tabular}
    \vspace{-1.25mm}
\caption{Metrics median values out of the $27$ users executions of the peg-in-hole task using the locomotion approach.}\label{loco_appr}
\vspace{-2.52mm}
\end{table}

\begin{table}[t!]
    \centering
    \begin{tabular}{@{}c@{\;\;\;\;\;\;\;\;\;\;\;\;\;\;}c@{\;\;\;\;\;\;\;\;\;\;\;\;\;\;}c@{\;\;\;\;\;\;\;\;\;\;\;\;\;\;}c@{\;\;\;\;\;\;\;\;\;\;\;\;\;\;}c@{}}
\toprule
        $\bar{E}\!$ [J] & $\bar{F}_h\!$ [N] & $\bar{v}\!$ $\left[\mbox{m}/\mbox{s}\right]$ & $x_f$ $\!$ [m] &  $T_f$ $\!$ [s] \\
\midrule
$0.95$ & $11.31$ & $0.14$ & $0.28$ & $32.97$ \\
\bottomrule
    \end{tabular}
    \vspace{-1.25mm}
\caption{Metrics median values out of the $27$ users executions of the peg-in-hole task using the switch approach.}\label{switch_appr}
\vspace{-2.52mm}
\end{table}

\begin{table}[t!]
    \centering
    \begin{tabular}{@{}c@{\;\;\;\;\;\;\;\;\;\;\;\;\;\;}c@{\;\;\;\;\;\;\;\;\;\;\;\;\;\;}c@{\;\;\;\;\;\;\;\;\;\;\;\;\;\;}c@{\;\;\;\;\;\;\;\;\;\;\;\;\;\;}c@{}}
\toprule
        $\bar{E}\!$ [J] & $\bar{F}_h\!$ [N] & $\bar{v}\!$ $\left[\mbox{m}/\mbox{s}\right]$ & $x_f$ $\!$ [m] &  $T_f$ $\!$ [s] \\
\midrule
$0.86$ & $12.30$ & $0.14$ & $0.08$ & $28.33$ \\
\bottomrule
    \end{tabular}
    \vspace{-1.25mm}
\caption{Metrics median values out of the $27$ users executions of peg-in-hole task using minimum energy approach.}\label{min_e_appr}
\vspace{-2.2mm} 
\end{table}

By comparing the median results in Table~\ref{loco_appr}, Table~\ref{switch_appr} and Table~\ref{min_e_appr}, the following conclusions can be drawn. The average human force $\bar{F}_h$ indicates that the locomotion approach requires the highest human effort due to the extensive use of the mobile base. In contrast, the switch and minimum-energy approaches reduce the effort by approximately $29\%$ and $22\%$, respectively. This is consistent with the corresponding differences in average execution speed $\bar{v}$ and with the mobile base being the heaviest system component. The mobile base usage is predominant in the locomotion approach, user-dependent in the switch approach (leading to higher cognitive load and slightly longer execution time), and automatically minimized in the minimum-energy approach through \textbf{Optimization Problem 1} in Sec.~\ref{prop_appr_optimi}.

\begin{table}[t]
\centering
\caption{Statistical significance analysis of the results reported in Fig.~\ref{fig:boxplots}.}
\vspace{-1.4mm}
\resizebox{\columnwidth}{!}{%
\label{tab:posthoc_results}
\begin{tabular}{l@{\;}lcl}
\hline
\textbf{Metric} & \textbf{Comparison} & \textbf{$p_{\mathrm{corr}}$} & \textbf{Significance} \\
\hline
$\bar{E}$ & Locomotion vs Switch & $1.682 \times 10^{-5}$ & Significant \\
$\bar{E}$ & Locomotion vs Minimum Energy & $1.682 \times 10^{-5}$ & Significant \\
$\bar{E}$ & Switch vs Minimum Energy & $0.337$ & Not significant \\
\hline
$\bar{F}_h\!$ & Locomotion vs Switch & $9.410 \times 10^{-4}$ & Significant \\
$\bar{F}_h\!$ & Locomotion vs Minimum Energy & $0.0153$ & Significant \\
$\bar{F}_h\!$ & Switch vs Minimum Energy & $0.0609$ & Not significant \\
\hline
$\bar{v}\!$ & Locomotion vs Switch & $0.0172$ & Significant \\
$\bar{v}\!$ & Locomotion vs Minimum Energy & $0.0977$ & Not significant \\
$\bar{v}\!$ & Switch vs Minimum Energy & $0.0977$ & Not significant \\
\hline
$x_f$ & Locomotion vs Switch & $1.682 \times 10^{-5}$ & Significant \\
$x_f$ & Locomotion vs Minimum Energy & $0.414$ & Not significant \\
$x_f$ & Switch vs Minimum Energy & $1.682 \times 10^{-5}$ & Significant \\
\hline
$T_f$ & Locomotion vs Switch & $0.0285$ & Significant \\
$T_f$ & Locomotion vs Minimum Energy & $0.0285$ & Significant \\
$T_f$ & Switch vs Minimum Energy & $4.412 \times 10^{-4}$ & Significant \\
\hline
\end{tabular}
}
\vspace{-1.4mm}
\end{table}

The average total kinetic energy $\bar{E}$ is approximately $62\%$ and $66\%$ lower in the switch and minimum-energy approaches, respectively, than in the locomotion approach. The switch and minimum-energy approaches achieve similar energy levels (as confirmed by the statistical non-significance of the corresponding metrics in Table~\ref{tab:posthoc_results}); however, the former relies on user intervention to limit mobile base usage, whereas the latter handles this trade-off automatically, resulting in a smoother energy profile. Moreover, in the switch approach the whole-body stored kinetic energy strongly depends on user experience in properly determining when to switch, while in the minimum-energy approach it is consistently regulated by \textbf{Optimization Problem 1}.

The absence of switching operations in locomotion and minimum-energy approaches also improves execution time, as shown by the metric $T_f$. In this case, the minimum-energy approach resulted to be the best performing one.

Regarding the final end-effector displacement $x_f$, the locomotion and minimum-energy approaches achieve comparable accuracy, with the latter performing slightly better. This expected result, confirmed by the statistical non-significance of the corresponding metric in Table~\ref{tab:posthoc_results}, is due to the fact that both exploit secondary tasks in the Jacobian null space for this purpose. In the switch approach, instead, the final displacement depends on user's action during manipulation, which may lead to suboptimal configurations.

\begin{figure}[t]
 \includegraphics[clip,width=0.959\linewidth]{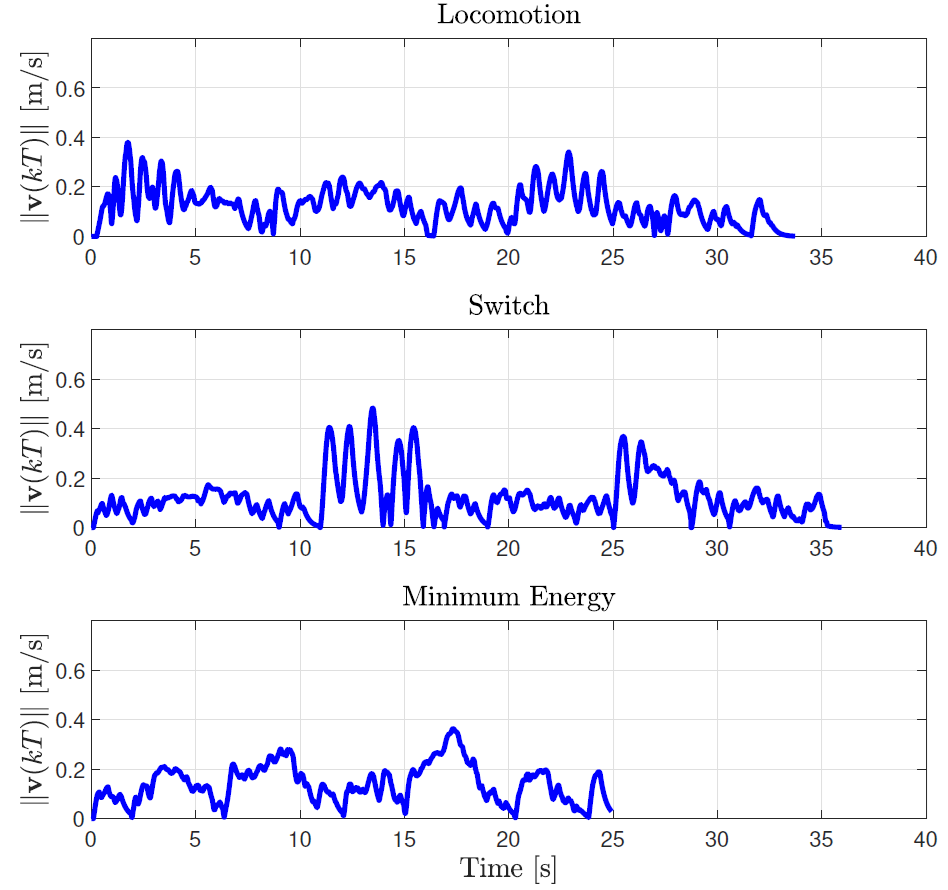}
    \vspace{-1.68mm}
    \caption{End-effector velocity during task execution by the user most frequently associated with median values in Fig.~\ref{fig:boxplots}.}
\label{figure_speed}
\vspace{-3.2mm} 
 \end{figure}

\begin{figure}[t]
 \includegraphics[clip,width=0.959\linewidth]{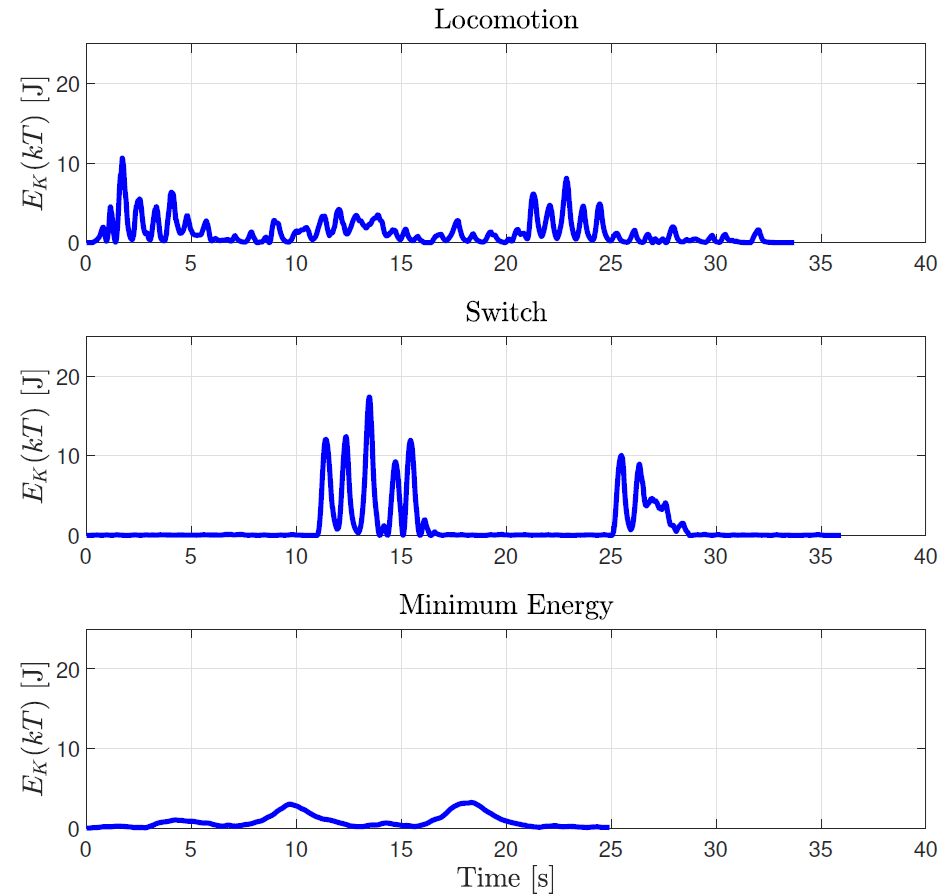}
    \vspace{-1.68mm}
    \caption{Total kinetic energy 
    during task execution by the user most frequently associated with median values in Fig.~\ref{fig:boxplots}.}
\label{figure_energy}
\vspace{-3.2mm} 
 \end{figure}

Fig.~\ref{figure_speed} and Fig.~\ref{figure_energy} show the norm $\left\|\boldsymbol{v}(kT)\right\|$ of the end-effector velocity and the total stored kinetic energy $E_K(kT)$ in \eqref{tot_kin_en} during task execution for the user most frequently associated with the median values in Fig.~\ref{fig:boxplots} and Tables~\ref{loco_appr}, \ref{switch_appr}, and \ref{min_e_appr}. From Fig.~\ref{figure_speed}, it can be observed that the locomotion approach exhibits more pronounced velocity fluctuations 
compared to the other approaches, resulting in a jagged and less stable motion profile. Since $\left\|\boldsymbol{v}(kT)\right\|$ is proportional to the momentum, this indicates more frequent and abrupt movements, leading to reduced user comfort and safety. At the same time, the switch approach shows similarly high peaks when the locomotion mode is activated, lower peaks when manipulation mode is activated. The minimum-energy approach represents the best compromise, as the use of the mobile base is automatically limited to strictly necessary movements in order to reduce the kinetic energy stored in the mobile base representing the heaviest system component, and the user does not need to handle the switching between different operating modes. 

Fig.~\ref{figure_energy} shows the time evolution of the total kinetic energy. In the locomotion approach, high energy peaks are observed due to large movements of the mobile base. In the switch approach, the stored energy is low during manipulation (i.e., when the manipulator arm is used) and increases significantly during locomotion phases (i.e., when the mobile base is used). Finally, the minimum-energy approach exhibits the smoothest profile with lower peak values, resulting in reduced overall kinetic energy stored in the whole-body system, thus enhancing human safety during the interaction and giving a smoother user experience.

\vspace{-2.102mm}
\section{Conclusions}\label{concl_sect}
\vspace{-1mm}

This paper addressed the control of a mobile manipulator for human–robot interaction, aiming to minimize the overall kinetic energy of the system. The approach was detailed and experimentally validated on a peg-in-hole task. Results show that the proposed minimum-energy method achieves the best overall performance across the selected metrics, leading to improved physical human–robot interaction. Future work will focus on task-dependent tuning of inertia-based weights to accommodate a wider range of tasks, and on enhancing user perception through haptic feedback to improve situational awareness for collision avoidance.

\bibliographystyle{IEEEtran}
\bibliography{bibliography}

\end{document}